\pdfoutput=1

\documentclass[11pt]{article}
\usepackage[dvipsnames]{xcolor}
\usepackage[final]{acl}

\usepackage{times}
\usepackage{latexsym}

\usepackage[T1]{fontenc}

\usepackage[utf8]{inputenc}

\usepackage{microtype}

\usepackage{inconsolata}

\usepackage{hyperref}
\usepackage{url}

\usepackage{graphicx}
\usepackage{subfigure}
\usepackage{wrapfig}

\usepackage{amsmath}
\usepackage{amstext}
\usepackage{amsfonts}
\usepackage{bm}

\usepackage{bbm}
\usepackage{multirow}
\usepackage{booktabs}
\usepackage{array}
\usepackage{caption}
\usepackage{multirow}
\usepackage{booktabs}
\usepackage{array}
\usepackage{caption}
\usepackage{color}
\usepackage{colortbl}
\usepackage{tablefootnote}
\usepackage{adjustbox}
\newcommand{\cred}[1]{\textcolor{red}{$_{#1}$}}

\usepackage[most]{tcolorbox}

\colorlet{lightSalmon}{Salmon!80}
\newcommand{\colorize}[2]{\colorbox{lightSalmon!#1!white}{\strut #2}}

\usepackage{caption}
\usepackage{algorithm}
\usepackage{algpseudocode}

\algnewcommand{\LineComment}[1]{\Statex ~~~~~~\textsc{//}~\textit{#1}}

\usepackage{enumitem}
\setenumerate[1]{itemsep=0pt,partopsep=0pt,parsep=\parskip,topsep=5pt}
\setitemize[1]{itemsep=0pt,partopsep=0pt,parsep=\parskip,topsep=5pt}
\setdescription{itemsep=0pt,partopsep=0pt,parsep=\parskip,topsep=5pt}

\usepackage{soul}

\usepackage{tikz}
\usepackage[edges]{forest}
\definecolor{hidden-draw}{RGB}{64,101,149}
\definecolor{hidden-pink}{RGB}{231,239,250}

\usepackage[mathscr]{euscript}
\newcommand{\M}{\mathcal{M}}

\usepackage{amssymb}  
\usepackage{pifont}

\usepackage{xspace}

\usepackage{twemojis}

\newcommand{\method}{\texttt{TokenSkip}\xspace}

\usepackage{ulem}

\title{
\texorpdfstring{\includegraphics[width=18pt]{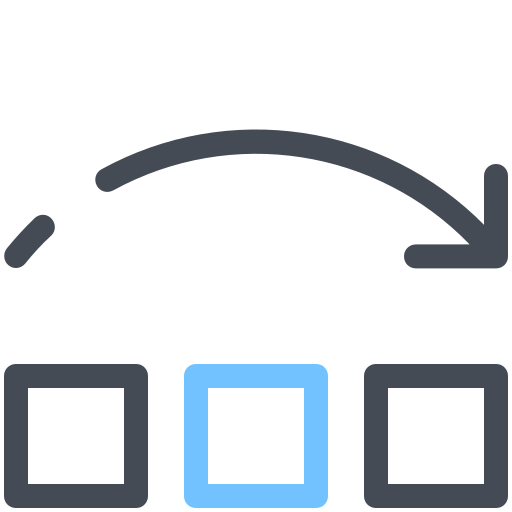}}{}
\method: Controllable Chain-of-Thought Compression in LLMs
}

\author{{Heming Xia}$^{\twemoji{peach}}$, {\textbf{Chak Tou Leong}}$^{\twemoji{peach}}$, {\textbf{Wenjie Wang}}$^{\twemoji{lemon}}$, {\textbf{Yongqi Li}}$^{\twemoji{peach}}$\thanks{Corresponding Author}, {Wenjie Li}$^{\twemoji{peach}}$\\
  $^{\twemoji{peach}}$ Department of Computing, The Hong Kong Polytechnic University \\
  $^{\twemoji{lemon}}$ University of Science and Technology of China \\
  {\tt \{he-ming.xia, chak-tou.leong\}@connect.polyu.hk}
}

\begin{document}
\maketitle
\def\thefootnote{\arabic{footnote}}
\begin{abstract}
Chain-of-Thought (CoT) has been proven effective in enhancing the reasoning capabilities of large language models (LLMs). Recent advancements, such as OpenAI's o1 and DeepSeek-R1, suggest that scaling up the length of CoT sequences during inference could further boost LLM reasoning performance. However, due to the autoregressive nature of LLM decoding, longer CoT outputs lead to a linear increase in inference latency, adversely affecting user experience, particularly when the CoT exceeds 10,000 tokens. To address this limitation, we analyze the semantic importance of tokens within CoT outputs and reveal that their contributions to reasoning vary. Building on this insight, we propose \method, a simple yet effective approach that enables LLMs to selectively skip less important tokens, allowing for controllable CoT compression. Extensive experiments across various models and tasks demonstrate the effectiveness of \method in reducing CoT token usage while preserving strong reasoning performance. Notably, when applied to Qwen2.5-14B-Instruct, \method reduces reasoning tokens by $40\%$ (from 313 to 181) on GSM8K, with less than a $0.4\%$ performance drop. We release our code and checkpoints in \url{https://github.com/hemingkx/TokenSkip}.
\end{abstract}

\section{Introduction}

Chain-of-Thought (CoT) prompting~\cite{Nye:2021, cot, Kojima:2022cotzero} has emerged as a cornerstone strategy for enhancing Large Language Models (LLMs) in complex reasoning tasks. By eliciting step-by-step inference, CoT enables LLMs to decompose intricate problems into manageable subtasks, thereby improving their problem-solving performance~\cite{Yao:2023tot, Wang:2023self-consistency, Zhou:2023least, Shinn:2023Reflexion}. Recent advancements, such as OpenAI's o1~\cite{o1} and DeepSeek-R1~\cite{deepseekr1}, further demonstrate that scaling up CoT lengths from hundreds to thousands of reasoning steps could continuously improve LLM reasoning. These breakthroughs have underscored CoT’s potential to advance LLM capabilities, expanding the boundaries of AI-driven problem-solving.

\begin{figure}[t]
\centering
    \includegraphics[width=0.95\columnwidth]{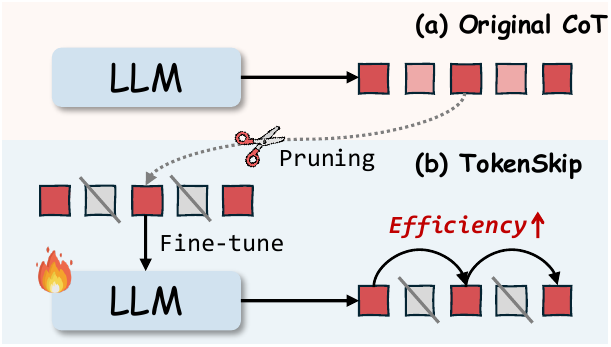}
    \caption{In contrast to vanilla CoT that generates all reasoning tokens sequentially, \method enables LLMs to \textit{skip} tokens with less semantic importance (\textit{e.g.,} \includegraphics[width=7pt]{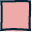}~) and learn shortcuts between critical reasoning tokens, facilitating controllable CoT compression.}
    \label{fig:intro}
\end{figure}

Despite its effectiveness, the increased length of CoT sequences introduces substantial computational overhead. Due to the autoregressive nature of LLM decoding, longer CoT outputs lead to proportional increases in both inference latency and memory footprints of key-value cache. Additionally, the quadratic computational cost of attention layers further exacerbates this burden. These issues become particularly pronounced when CoT sequences extend into thousands of reasoning steps, resulting in significant computational costs and prolonged response times. While prior research has explored methods for selectively skipping reasoning steps~\cite{Ding:2024cotshortcut, liu2024skipstep}, recent findings~\cite{jin:2024cotlength, Merrill:2024cotlength} suggest that such reductions may conflict with test-time scaling~\cite{o1-blog, snell2025scaling}, ultimately impairing LLM reasoning performance. Therefore, striking an optimal balance between CoT efficiency and reasoning accuracy remains a critical open challenge.

In this work, we delve into CoT efficiency and seek the answer to an important question: \textit{``Does every token in the CoT output contribute equally to deriving the answer?''} We empirically analyze the semantic importance of tokens within CoT outputs and reveal that their contributions to the reasoning performance vary, as depicted in Figure 2. Building on this insight, we introduce \method, a simple yet effective approach that enables LLMs to \textit{skip} less important tokens within CoT sequences and learn shortcuts between critical reasoning tokens, thereby allowing for controllable CoT compression with adjustable ratios. Specifically, as shown in Figure~\ref{fig:intro}, \method constructs compressed CoT training data with various compression ratios, by pruning unimportant tokens from original LLM CoT trajectories. Then, it conducts a general supervised fine-tuning process on target LLMs with this training data, facilitating LLMs to automatically trim redundant tokens during reasoning.

We conduct extensive experiments across various models, including LLaMA-3.1-8B-Instruct and the Qwen2.5-Instruct series, using two widely recognized math reasoning benchmarks: GSM8K and MATH-500. The results validate the effectiveness of \method in compressing CoT outputs while maintaining robust reasoning performance. Notably, Qwen2.5-14B-Instruct exhibits almost \textbf{NO} performance drop (less than $0.4\%$) with a $\bm{40\%}$ reduction in token usage on GSM8K. On the challenging MATH-500 dataset, LLaMA-3.1-8B-Instruct effectively reduces CoT token usage by $\bm{30}\%$ with a performance decline of less than $4\%$, resulting in a $\bm{1.4}\times$ inference speedup. Further analysis underscores the coherence of \method in specified compression ratios and its potential scalability with stronger compression techniques.

\method is distinguished by its low training cost. For Qwen2.5-14B-Instruct, \method fine-tunes only 0.2\% of the model's parameters using LoRA. The size of the compressed CoT training data is no larger than that of the original training set, with 7,473 examples in GSM8K and 7,500 in MATH. The training is completed in approximately 2 hours for the 7B model and 2.5 hours for the 14B model on two 3090 GPUs. These characteristics make \method an efficient and reproducible approach, suitable for use in efficient and cost-effective LLM deployment.

To sum up, our key contributions are:
\begin{enumerate}
    \item To the best of our knowledge, this work is the \textit{first} to investigate the potential of enhancing CoT efficiency through \textit{token skipping}, inspired by the varying semantic importance of tokens in CoT trajectories of LLMs.
    \item We introduce \method, a simple yet effective approach that enables LLMs to skip redundant tokens within CoTs and learn shortcuts between critical tokens, facilitating CoT compression with adjustable ratios.
    \item Our experiments validate the effectiveness of \method. When applied to Qwen2.5-14B-Instruct, \method reduces reasoning tokens by $40\%$ (from 313 to 181) on GSM8K, with less than a $0.4\%$ performance drop.
\end{enumerate}

\section{Background and Preliminaries}
\label{sec:preliminaries}

In this section, we discuss the relevant research background and present preliminary studies on token efficiency in CoT sequences, exploring its impact on the reasoning performance of LLMs.

\subsection{Token Importance}
\label{sec:token-importance}

We first investigate a critical research question to CoT efficiency: \textit{``Does every token in the CoT output contribute equally to deriving the answer?''} In other words, we would like to know if there is any token redundancy in CoT sequences that could be eliminated to improve CoT efficiency.

Token redundancy has been recognized as a longstanding and fundamental issue in LLM efficiency~\cite{hou:2022tokendropbert, zhang2023h2o, lin2024criticaltokenpretrain, Chen:2024FastV}. Recently, it has garnered intensive research attention in prompt compression~\cite{li:2023selective, jiang2023:llmlingua, pan:2024llmlingua2}, which focuses on removing redundant tokens from the input prompt to reduce API token usage. To address this issue, Selective Context~\cite{li:2023selective} proposed to measure the importance of tokens in a piece of text based on the semantic confidence of LLMs:
\begin{equation}
I_1\left(x_i\right)=-\log P\left(x_i \mid \bm{x}_{<{i}}; \bm{\theta}_{\M_L}\right),
\label{eq:selectivecontext}
\end{equation}
where $\boldsymbol{x}=\left\{x_i\right\}_{i=1}^{n}$ is the given text, $x_i$ denotes a token, and $\M_L$ denotes the LLM used to compute the confidence of each token. Intuitively, such a measurement could be seamlessly applied to CoT tokens generated by LLMs. We show an example of this measurement in Figure~\ref{fig:token-importance}.

\begin{figure}[t]
    \centering
    \resizebox{\columnwidth}{!}{
    \fbox{\parbox[c]{1.1\columnwidth}{
        \textbf{Problem: } Marcus is half of Leo’s age and five years younger than Deanna. Deanna is 26. How old is Leo?

        \vskip 0.1in

        \textbf{Chain-of-Thought: } {\setlength{\fboxsep}{-1pt}
         \input{inputs/text_file/token_importance_llama}
        }

        \vskip 0.1in

        \textbf{Chain-of-Thought: } {\setlength{\fboxsep}{-1pt}
         \input{inputs/text_file/token_importance_llmlingua}
        }

        \vskip 0.05in

        \textbf{Final Answer: } 42.
    }}}
    \caption{Visualization of token importance within a CoT sequence, with darker colors indicating higher values. This figure compares two token importance measurements: Selective Context and LLMLingua-2.}
    \label{fig:token-importance}
\end{figure}

Despite its simplicity, LLMLingua-2~\cite{pan:2024llmlingua2} argued that there exist two major limitations in the aforementioned measurement that hinder the compression performance. Firstly, as shown in Figure~\ref{fig:token-importance}, the intrinsic nature of LLM perplexity leads to lower importance measures (i.e., higher confidence) for tokens at the end of the sentence. Such position dependency impacts the factual importance measurement of each token. Furthermore, the unidirectional attention mechanism in causal LMs may fail to capture all essential information needed for token importance within the text. 

To tackle these limitations, LLMLingua-2 introduced utilizing a bidirectional BERT-like LM~\cite{bert} for token importance measurement. It utilizes GPT-4~\cite{gpt-4} to label each token as ``\textit{important}'' or not and trains the bidirectional LM with a token
classification objective. The token importance is measured by the predicted probability of each token:
\begin{equation}
I_2\left(x_i\right)= P\left(x_i \mid \bm{x}_{\le n}; \bm{\theta}_{\M_B}\right),
\label{eq:llmlingua2}
\end{equation}
where $\M_B$ denotes the bidirectional LM. 

This study applies LLMLingua-2 as the importance measurement to CoT tokens. Similar to plain text, we observe that the semantic importance of tokens within CoT outputs varies, as shown in Figure~\ref{fig:token-importance}. For instance, mathematical equations tend to have a greater contribution to the final answer, consistent with recent research~\cite{Ma:2024mathmatters}. In contrast, semantic connectors such as ``\textit{so}'' and ``\textit{since}'' generally contribute less. These findings highlight the token redundancy in CoT outputs of LLMs and the substantial potential to enhance CoT efficiency by trimming this redundancy.

\begin{figure}[t]
\begin{tcolorbox}[colback=blue!5!white,colframe=blue!75!black,title=Revovering the Compressed Chain-of-Thought,fontupper=\footnotesize,fonttitle=\scriptsize]
\textbf{Compressed CoT}: break down Deanna 26 Marcus five younger 26 - 5 21 Marcus half Leo's age twice Marcus Marcus 21, Leo's age 2 x 21 = 42.

\vskip 0.1in
        
\textbf{Recovered Compressed CoT}: Let's break it down step by step. Deanna is 26 years old. Marcus is five years younger than Deanna: M = D - 5. Marcus's age: M = 26 - 5 = 21. Marcus is half of Leo's age: M = L / 2. Leo is twice Marcus's age: L = 2M. Leo's age: L = 2 x 21 = 42.

\end{tcolorbox}
\caption{Recovering the compressed CoT for GSM8K math word problem using LLaMA-3.1-8B-Instruct.}
\label{fig:recovery}
\end{figure}

\subsection{CoT Recovery}
\label{sec:cot-recovery}
We further explore the following research question: \textit{``Are LLMs capable of restoring the CoT process from compressed outputs?''} The answer is yes. As shown in Figure~\ref{fig:recovery} and detailed in Appendix~\ref{appendix:recovery}, examples restored from compressed CoTs using LLaMA-3.1-8B-Instruct demonstrate that LLMs could effectively comprehend the semantic information encoded in the compressed CoT and restore the CoT process. This capability ensures that the interpretability of compressed CoTs is maintained. Additionally, when required by users, the complete CoT process can be recovered and presented.

In summary, the empirical analysis above underscores the potential of trimming redundant tokens to enhance CoT efficiency, as well as the ability of LLMs to restore CoT from compressed outputs. However, enabling LLMs to autonomously skip redundant CoT tokens and identify shortcuts between critical reasoning tokens presents a non-trivial challenge. To the best of our knowledge, this work is the \textit{first} to explore CoT compression through \textit{token skipping}. In the following sections, we present our proposed methodology in detail.

\section{\method}
\label{sec:tokenskip}
\begin{figure*}[t]
\centering
\includegraphics[width=0.95\textwidth]{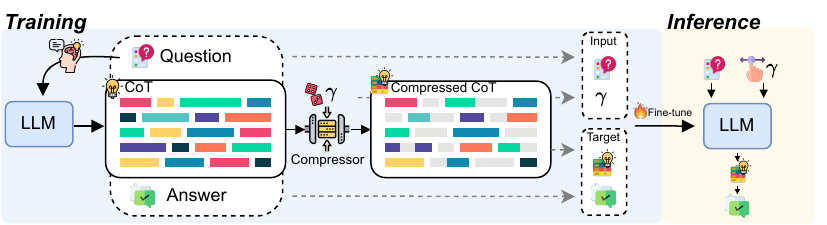}
\caption{Illustration of \method. During training, \method first generates CoT trajectories from the target LLM. These CoTs are then compressed to various ratios sampled from the ratio set. \method fine-tunes the LLM using compressed CoTs with mixed ratios, enabling controllable CoT inference at any desired $\gamma \in \left\{\gamma_0,\dots,\gamma_z\right\}$.}
\label{fig:tokenskip}
\end{figure*}

We introduce \method, a simple yet effective approach that enables LLMs to skip less important tokens, enabling controllable CoT compression with adjustable ratios. This section demonstrates the details of our methodology, including token pruning~(\S\ref{sec:token-pruning}), training~(\S\ref{sec:training}), and inference~(\S\ref{sec:inference}).

\subsection{Token Pruning}
\label{sec:token-pruning}
The key insight behind \method is that ``\textit{each reasoning token contributes differently to deriving the answer.}'' To enhance CoT efficiency, we propose to trim redundant CoT tokens from LLM outputs and fine-tune LLMs using these trimmed CoT trajectories. The token pruning process is guided by the concept of \textit{token importance}, as detailed in Section~\ref{sec:token-importance}. 

Specifically, given a target LLM $\M$, one of its CoT trajectories $\boldsymbol{c}=\left\{c_i\right\}_{i=1}^{m}$, and a specified compression ratio $\gamma \in \left[0,1\right]$ for the current $\boldsymbol{c}$, \method first calculates the semantic importance of each CoT token $\left\{I(c_{i})\right\}_{i=1}^{m}$, as defined in Eq~(\ref{eq:llmlingua2}), and then ranks the resulting scores in descending order. The empirical $\gamma$-quantile of these importance values serves as the pruning threshold:
\begin{equation}
I_\gamma=Q_{\gamma}\left(I\left(c_1\right), . ., I\left(c_m\right)\right),
\end{equation}
where $Q_{\gamma}$ denotes the $\gamma$-quantile (i.e.\ the $\gamma$-th percentile) of the multiset $\left\{I(c_{i})\right\}_{i=1}^{m}$. All CoT tokens whose importance value meets or exceeds this threshold are retained, yielding the compressed CoT trajectory:
\begin{equation}
\widetilde{\boldsymbol{c}}=\left\{c_i \mid I\left(c_i\right) \geq I_\gamma, 1 \leq i \leq m\right\}.
\end{equation}

\subsection{Training}
\label{sec:training}
Given a training dataset $\mathcal{D}$ with $N$ samples and a target LLM $\M$, we first obtain $N$ CoT trajectories with $\M$. Then, we filter out trajectories with incorrect answers to ensure data quality. For the remaining trajectories, we prune each CoT with a compression ratio $\gamma$ sampled from the ratio set $\left\{\gamma_0,\dots,\gamma_z\right\}$, as demonstrated in Section~\ref{sec:token-pruning}. For each $\langle\text{question}, \text{compressed CoT}, \text{answer}\rangle$, we inserted the compression ratio $\gamma$ after the question. Each training sample is formatted as follows: 
\begin{equation}
\nonumber
    \mathcal{Q} \ \mathrm{[EOS]} \ \gamma \ \mathrm{[EOS]} \ \mathrm{Compressed\ CoT} \ \mathcal{A},
\end{equation}
where $\langle\mathcal{Q}, \mathcal{A}\rangle$ indicates the $\langle\text{question}, \text{answer}\rangle$ pair. Formally, given a question $\boldsymbol{x}$, a compression ratio $\gamma$ randomly sampled from $\left\{\gamma_0,\dots,\gamma_z\right\}$, and the output sequence $\boldsymbol{y}=\left\{y_i\right\}_{i=1}^{l}$, which includes the compressed CoT $\widetilde{\boldsymbol{c}}$ and the answer $\boldsymbol{a}$, we fine-tunes the target LLM $\M$, enabling it to perform chain-of-thought in a compressed pattern by minimizing
\begin{equation}
\mathcal{L}=\sum_{i=1}^{l} \log P\left(y_{i} \mid \bm{x}, \gamma, \bm{y}_{<i}; \bm{\theta}_{\M}\right),
\end{equation}
where $\bm{y} =\left\{\widetilde{c}_1, \cdots,\widetilde{c}_{m^{\prime}}, a_1, \cdots, a_t  \right\}$. Note that the compression is performed solely on CoT sequences, and we keep the answer $\boldsymbol{a}=\left\{a_i\right\}_{i=1}^{t}$ unchanged. To preserve LLMs' reasoning capabilities, we also include a portion of the original CoT trajectories in the training data, with $\gamma$ set to 1.

\subsection{Inference}
\label{sec:inference}
The inference of \method follows autoregressive decoding. Compared to original CoT outputs that may contain redundancy, \method facilitates LLMs to skip \textit{unimportant} CoT tokens, thereby enhancing reasoning efficiency. Formally, given a question $\boldsymbol{x}$ and a desired compression ratio $\gamma \in \left\{\gamma_0,\dots,\gamma_z\right\}$, the input prompt of \method follows the same format adopted in fine-tuning, which is $\mathcal{Q} \ \mathrm{[EOS]} \ \gamma \ \mathrm{[EOS]}$. The LLM $\M$ sequentially predicts the output sequence $\hat{\bm{y}}$:
\begin{equation}
\nonumber
\hat{\boldsymbol{y}}=\arg \max _{\boldsymbol{y}^*} \sum_{j=1}^{l^{\prime}} \log P\left(y_j \mid \boldsymbol{x}, \gamma, \boldsymbol{y}_{<j}; \bm{\theta}_{\M}\right),
\end{equation}
where $\hat{\bm{y}} =\left\{\hat{c}_1, \cdots,\hat{c}_{m^{\prime\prime}}, \hat{a}_1, \cdots, \hat{a}_{t^{\prime}}  \right\}$ denotes the output sequence, which includes CoT tokens $\hat{\bm{c}}$ and the answer $\bm{\hat{a}}$. We illustrate the training and inference process of \method in Figure~\ref{fig:tokenskip}. 

\section{Experiments}
\subsection{Experimental Setup}
\label{sec:exp_setup}

\paragraph{Models and Datasets} 
We primarily evaluate our method using LLaMA-3.1-8B-Instruct~\cite{dubey:2024llama3} and Qwen2.5-Instruct series~\cite{Qwen2}. The evaluation leverages two widely-used math reasoning benchmarks: GSM8K~\citep{Cobbe:2021gsm8k} and MATH~\cite{math}. For training, we use the respective training sets from both datasets. Regarding the MATH dataset, due to the computation cost, we assess our method on a subset, MATH-500, which is identical to the test set used in \citet{Lightman:2024verify}.

\paragraph{Implementation Details} 
We utilize LLMLingua-2~\cite{pan:2024llmlingua2} as the token importance metric to generate our compressed CoT training data. The compression ratio $\gamma$ is randomly selected from the ratio set $\{0.5, 0.6, 0.7, 0.8, 0.9, 1.0\}$ for each training sample. We adopt LoRA~\cite{lora} to train our models. \method is characterized by its low training cost, with training taking $\sim$2 hours for the 7B model and $\sim$2.5 hours for the 14B model on 3090 GPUs. We include more implementation details in Appendix~\ref{appendix:training_details}.

\paragraph{Baselines} 
We compare \method to three baselines: \textbf{1) Token-efficient Prompts.} Following \citet{Lee:2025Complexity}, we select three advanced prompts, instructing LLMs to perform CoT efficiently. These prompts, denoted as \texttt{BeConcise}, \texttt{OnlyNumbers}, and \texttt{AbbreWords}, are detailed in Appendix~\ref{appendix:prompt_detail}; \textbf{2) Length-control Prompts.} We instruct the LLM to reduce a fixed proportion of output tokens in the CoT process, denoted as \texttt{LC-Prompt} in Table~\ref{tab:main}; \textbf{3) Truncation.} This method involves brute-force length truncation, where the maximum number of output tokens is restricted, compressing the CoT output to a fixed ratio.

\begin{figure}[t]
\centering
    \includegraphics[width=0.95\columnwidth]{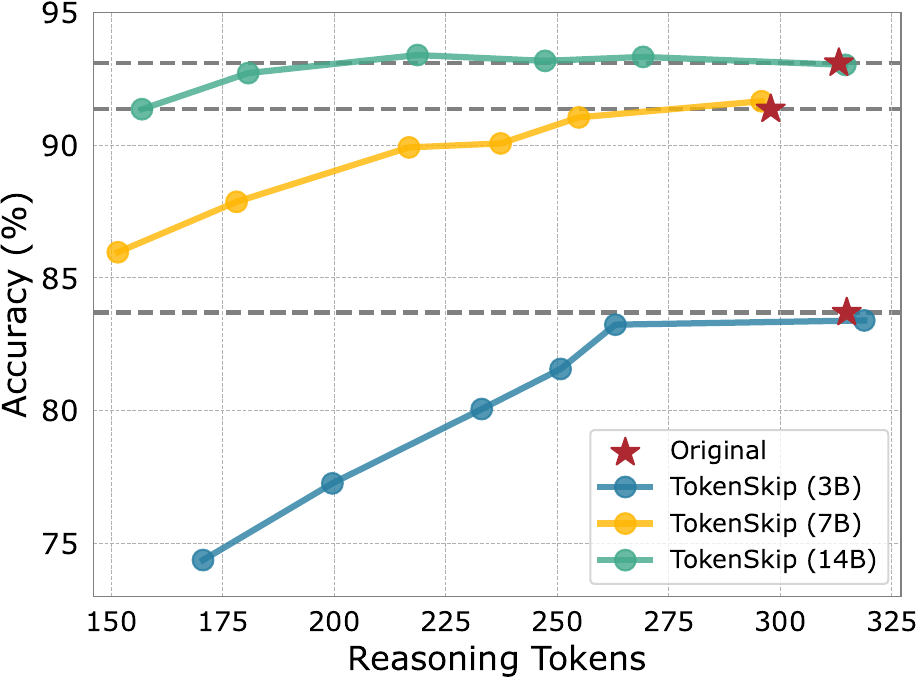}
    \caption{Compression performance of \method on Qwen2.5-Instruct models. Qwen2.5-14B-Instruct shows almost \textbf{no} performance drop with $\bm{40\%}$ token trimming.}
    \label{fig:qwen-scaling}
\end{figure}

\paragraph{Evaluation Metrics} 
We evaluate \method using three widely used metrics: accuracy, the number of CoT tokens, and inference latency per sample. Model performance is assessed using scripts from \texttt{DeepSeek-Math}\footnote{\url{https://github.com/deepseek-ai/DeepSeek-Math}}. Greedy decoding is employed to generate the outputs from the target LLM. Inference latency is measured on a single NVIDIA 3090 GPU with a batch size of 1. In addition to these metrics, we report the actual compression ratio of the CoTs to assess whether the compression aligns with the specified ratio.

\begin{table*}[t]
\centering
\small
\setlength{\tabcolsep}{1.4mm}
\begin{tabular}{@{}lcrrrcrrrc@{}}
\toprule
\multirow{2}{*}{\textbf{Methods}} &\multirow{2}{*}{$\textbf{Ratio}$} & \multicolumn{4}{c}{\textbf{GSM8K}}  & \multicolumn{4}{c}{\textbf{MATH-500}} \\ \cmidrule(lr){3-6} \cmidrule(lr){7-10}
&  &Accuracy $\uparrow$ &Tokens $\downarrow$ &Latency (s) $\downarrow$ &\textit{Act}Ratio   &Accuracy $\uparrow$ &Tokens $\downarrow$ &Latency (s) $\downarrow$ &\textit{Act}Ratio  \\ \midrule
\texttt{Original}  & - & 86.2\cred{(0.0 \downarrow)} &213.17 &5.96\cred{1.0\times} &-  & 48.6\cred{(0.0 \downarrow)} & 502.60  &16.37\cred{1.0\times} &- \\ \midrule
\texttt{BeConcise}  & - & 82.9\cred{(3.3 \downarrow)} &161.32 &4.73\cred{1.3\times} &0.76  & 47.4\cred{(1.2 \downarrow)} & 471.34  &15.54\cred{1.1\times} &0.94 \\
\texttt{OnlyNumbers}  & - & 83.2\cred{(3.0 \downarrow)} &165.27 &4.95\cred{1.2\times} &0.78  & 46.4\cred{(2.2 \downarrow)} & 487.00  &15.93\cred{1.0\times} &0.97 \\
\texttt{AbbreWords}  & - & 83.7\cred{(2.5 \downarrow)} &170.33 &5.15\cred{1.2\times} &0.80  & 47.6\cred{(1.0 \downarrow)} & 489.07  &15.94\cred{1.0\times} &0.97 \\ \midrule
\multirow{3}{*}{\texttt{LC-Prompt}}  &0.9 & 84.1\cred{(2.1 \downarrow)} &226.37 &6.12\cred{1.0\times} & 1.06  & 48.6\cred{(0.0 \downarrow)} &468.04 &15.39\cred{1.1\times} & 0.93  \\
&0.7 & 84.9\cred{(1.3 \downarrow)} &209.39 &5.51\cred{1.1\times} & 0.98  & 48.4\cred{(0.4 \downarrow)} &472.13 &15.55\cred{1.1\times} & 0.94  \\
&0.5 & 83.7\cred{(2.5 \downarrow)} &188.82 &4.97\cred{1.2\times} & 0.89  & 47.8\cred{(0.4 \downarrow)} &471.11 &15.48\cred{1.1\times} & 0.94  \\ \midrule
\multirow{3}{*}{\texttt{Truncation}}  &0.9 & 70.2\cred{(26.0 \downarrow)} &202.06 &5.29\cred{1.1\times} & 0.95  & 47.8\cred{(0.8 \downarrow)} &440.33 &14.56\cred{1.1\times} & 0.88  \\
&0.7 & 25.9\cred{(60.3 \downarrow)} &149.99 &3.97\cred{1.5\times} & 0.70  & 45.0\cred{(3.6 \downarrow)} &386.89 &12.85\cred{1.3\times} & 0.77  \\
&0.5 & 7.0\cred{(79.2 \downarrow)} &103.69 &2.95\cred{2.0\times} & 0.49  & 27.4\cred{(21.2 \downarrow)} &283.70 &9.40\cred{1.7\times} & 0.56  \\ \midrule
\multirow{6}{*}{\method}  &1.0 & 86.7\cred{(0.5 \uparrow)} &213.60 &5.98\cred{1.0\times} & 1.00  & 48.2\cred{(0.4 \downarrow)} &504.79 &16.43\cred{1.0\times} & 1.00  \\
&0.9 & 86.1\cred{(0.1 \downarrow)} &198.01 &5.65\cred{1.1\times} & 0.93  & 47.8\cred{(0.8 \downarrow)} & 448.31  & 15.26\cred{1.1\times} & 0.89 \\
&0.8 & 84.3\cred{(1.9 \downarrow)} &169.89 &5.13\cred{1.2\times} & 0.80  & 47.3\cred{(1.3 \downarrow)} & 398.94  & 13.39\cred{1.2\times} & 0.79 \\
&0.7 & 82.5\cred{(3.7 \downarrow)} &150.12 &4.36\cred{1.4\times} & 0.70  & 46.7\cred{(1.9 \downarrow)} & 349.13  & 11.55\cred{1.4\times} & 0.69 \\
&0.6 & 81.1\cred{(5.1 \downarrow)} &129.38 &3.81\cred{1.6\times} & 0.61  & 42.0\cred{(6.6 \downarrow)} & 318.36  & 10.58\cred{1.6\times} & 0.63 \\
&0.5 & 78.2\cred{(8.0 \downarrow)} &113.05 &3.40\cred{1.8\times} & 0.53  & 40.2\cred{(8.4 \downarrow)} & 292.17  & 9.67\cred{1.7\times} & 0.58 \\
\bottomrule
\end{tabular}
\caption{Experimental results of \method on LLaMA-3.1-8B-Instruct. We report accuracy, average CoT token count (Tokens), average latency per sample, and actual compression ratio (\textit{Act}Ratio) for comparison.}
\label{tab:main}
\end{table*}

\subsection{Main Results}
\label{sec:main-exp}
The performance of \method on GSM8K using the Qwen2.5-Instruct series\footnote{For detailed results, please refer to Appendix~\ref{appendix:qwen_detail}.} is illustrated in Figure~\ref{fig:qwen-scaling}. As the model scale increases, there is less performance degradation at higher compression ratios, indicating that larger LLMs are better at identifying shortcuts between critical reasoning tokens, enabling more efficient CoT generation. Notably, Qwen2.5-14B-Instruct exhibits almost \textbf{NO} performance drop (less than $0.4\%$) with $\bm{40\%}$ token trimming. Even at a compression ratio of 0.5, the model maintains strong reasoning capabilities, with only $2\%$ performance degradation. These results highlight the substantial potential of \method to reduce CoT token usage and accelerate reasoning in large-scale LLMs.

Table~\ref{tab:main} compares \method with three widely used baselines. As shown, prompting methods, including token-efficient prompts and length-control ones, fail to achieve desired compression ratios. Specifically, token-efficient prompts achieve only 0.94-0.97 compression ratios on MATH-500, with nearly no efficiency improvements; the actual ratio of \texttt{LC-Prompt} exceeds 0.89 even when the target is set to 0.5. While \texttt{Truncation} adheres to the specified ratio, it results in significant degradation in reasoning performance. Concretely, at a compression ratio of 0.5, \texttt{Truncation} causes a $79\%$ accuracy drop on GSM8K and a $21\%$ drop on MATH-500. In contrast, \method ensures adherence to various desired compression ratios (see Figure~\ref{fig:allratio}) while preserving strong reasoning capabilities. Notably, \method achieves an actual compression ratio of \textbf{0.53} on GSM8K with merely a $10\%$ performance drop, resulting in a $\bm{1.8}\times$ speedup in average latency. On MATH-500, \method effectively reduces CoT token usage by $\bm{30}\%$ with a performance drop of less than $4\%$. These results validate the effectiveness of \method.

In Appendix~\ref{appendix:exp}, we illustrate additional experiments to evaluate the out-of-domain performance of \method and validate its generalizability beyond mathematical reasoning.
\subsection{Analysis}
\label{sec:analysis}
\begin{figure}[t]
\centering
    \includegraphics[width=0.95\columnwidth]{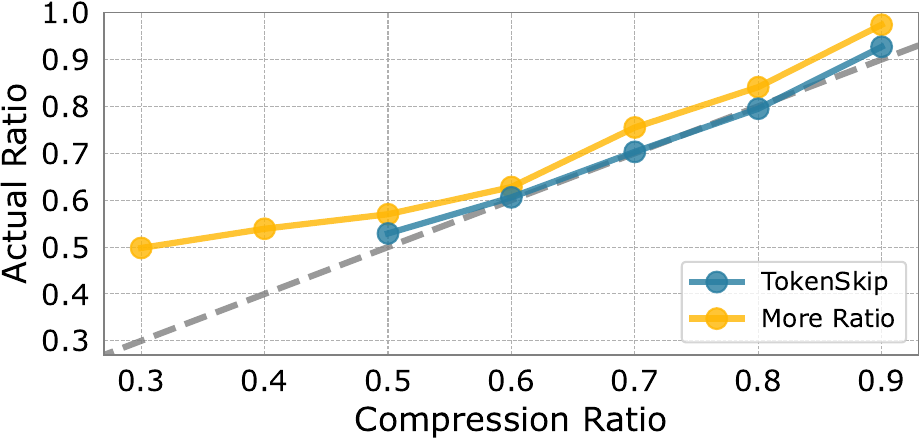}
    \caption{Comparison of ratio adherence across different compression ratio settings. The experimental results are obtained with LLaMA-3.1-8B-Instruct on GSM8K.}
    \label{fig:allratio}
\end{figure}

\begin{figure}[t]
\centering
    \includegraphics[width=0.95\columnwidth]{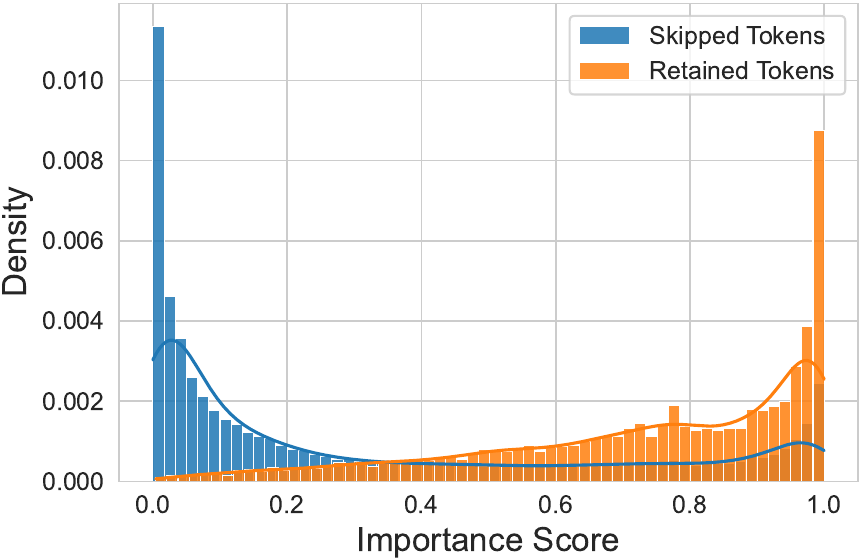}
    \caption{Distribution of token importance for skipped versus retained tokens. The LLM effectively learns to skip low-importance tokens and retain critical ones.}
    \label{fig:importance_dist}
\end{figure}

\paragraph{Compression Ratio} 
In our main results, we focus on compression ratios greater than 0.5. To further investigate the performance of \method at lower compression ratios, we train an additional variant, denoted as \texttt{More Ratio}, with extra compression ratios of 0.3 and 0.4. As shown in Figure~\ref{fig:allratio}, the ratio adherence of models largely degrades at these lower ratios. We attribute this decline to the excessive trimming of reasoning tokens, which likely causes a loss of critical information in the completions, hindering the effective training of LLMs to learn CoT compression. Furthermore, we observe that the overall adherence of \texttt{More Ratio} is not as good as \method with the default settings, which further supports our hypothesis.

\paragraph{Importance Distribution} To validate that the LLM learns to skip less important tokens, we analyzed the distribution of the number of tokens with various token importance. Specifically, we instructed \method with Qwen2.5-14B-Instruct to generate full CoTs ($\gamma=1.0$) and compressed CoTs ($\gamma=0.7$) on the GSM8K test set. CoT Tokens appearing exclusively in full CoTs but not in compressed ones were identified as ``\textit{skipped}'' while those present in compressed CoTs were considered ``\textit{retained}''. As illustrated in Figure~\ref{fig:importance_dist}, the importance distribution of skipped tokens skews towards lower values, whereas retained tokens predominantly exhibit higher importance. This demonstrates that \method effectively enables LLMs to discard less critical CoT tokens during inference.

\begin{figure}[t]
\centering
    \includegraphics[width=0.95\columnwidth]{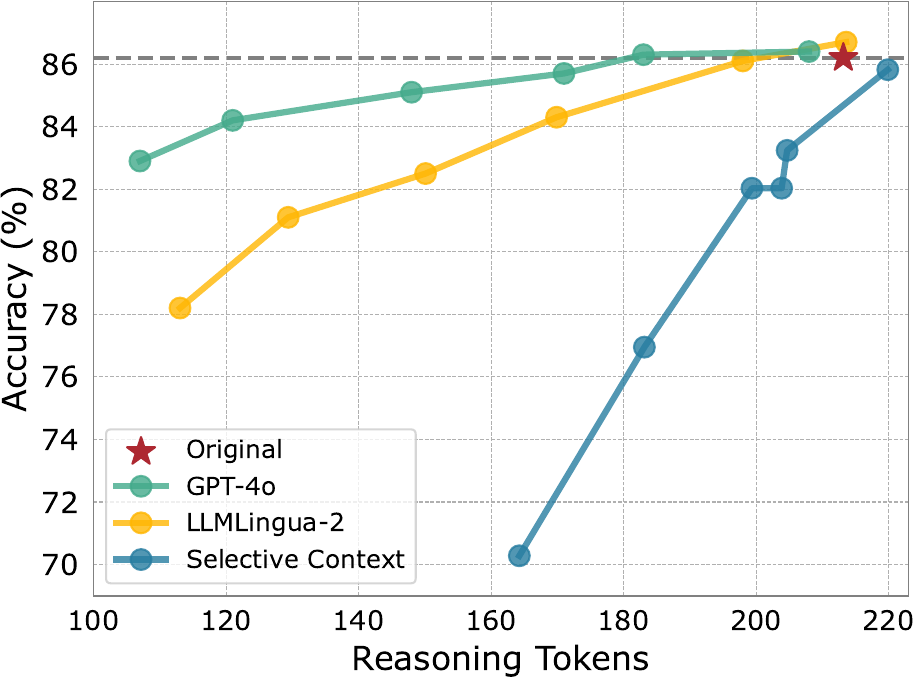}
    \caption{Performance comparison of \method using different token importance metrics, evaluated with LLaMA-3.1-8B-Instruct on GSM8K.}
    \label{fig:compressor}
\end{figure}

\paragraph{Importance Metric} 
Figure~\ref{fig:compressor} presents a comparison of \method across different importance metrics. In addition to the metrics discussed in Section \ref{sec:token-importance}, we include \texttt{GPT-4o}\footnote{We use the \texttt{gpt-4o-2024-08-06} version for experiments.} as a token importance upperbound for comparison. Specifically, for a given CoT trajectory, we prompt \texttt{GPT-4o} to trim redundant tokens according to a specified compression ratio, without adding any additional tokens. As shown in Figure~\ref{fig:compressor}, \method utilizing LLMLingua-2~\cite{pan:2024llmlingua2} outperforms the variant with Selective Context~\cite{li:2023selective}, which aligns with our demonstrations in Section \ref{sec:token-importance}. Additionally, the results of \texttt{GPT-4o} suggest that the capabilities of effective token importance metrics (beyond LLMLingua-2) could be further improved. However, the API costs associated with \texttt{GPT-4o} make it impractical for processing large-scale datasets. In contrast, LLMLingua-2, which includes a BERT-size model, offers a cost-effective and efficient alternative for training \method.

\begin{figure}[t]
\centering
    \includegraphics[width=0.95\columnwidth]{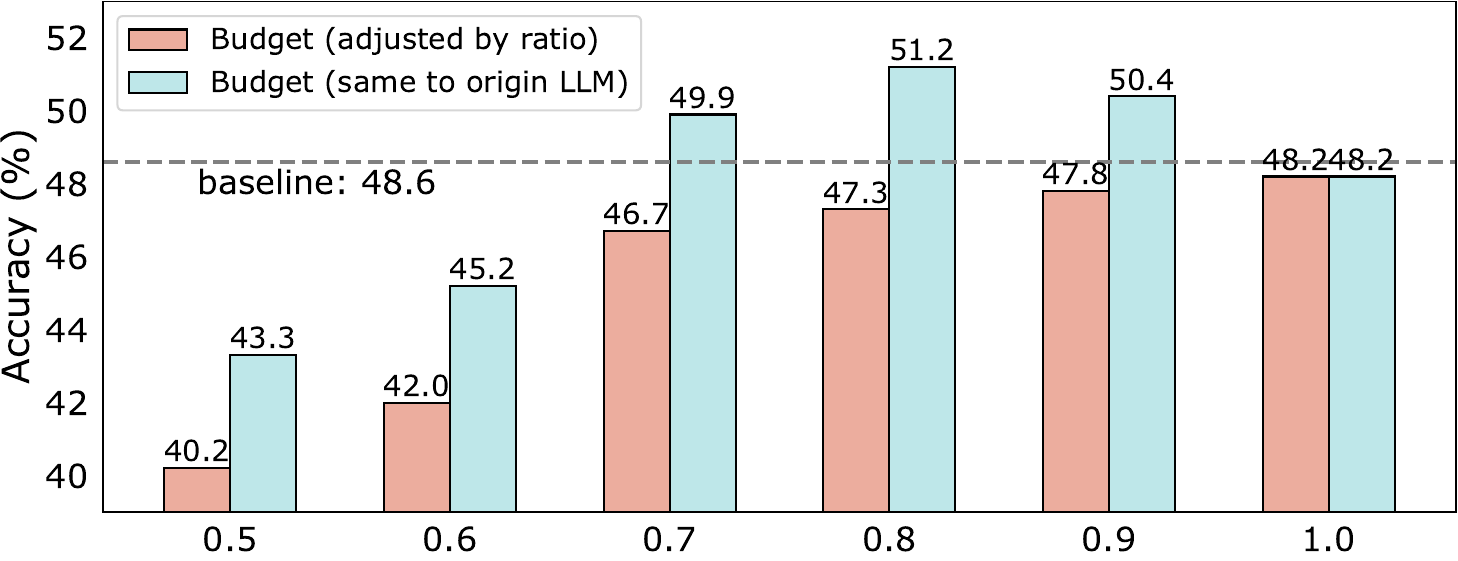}
    \caption{Performance comparison of \method with varying maximum length constraints, evaluated with LLaMA-3.1-8B-Instruct on the MATH-500 dataset.}
    \label{fig:budget}
\end{figure}

\paragraph{Length Budget} 
As outlined in Section~\ref{sec:exp_setup}, we adjust the maximum length budget to \texttt{max\_len}$\times\gamma$ when evaluating \method on MATH-500, ensuring a fair comparison of compression ratios. However, this brute-force length truncation inevitably impacts the reasoning performance of LLMs, as LLMs are unable to complete the full generation. In this analysis, we explore whether LLMs can ``\textit{think}'' more effectively using a compressed CoT format. Specifically, we evaluate \method under the same length budget as the original LLM (e.g., 1024 for MATH-500). The experimental results, shown in Figure~\ref{fig:budget}, demonstrate a significant performance improvement of \method under this length budget, compared to those adjusted by compression ratios. Notably, with compression ratios of 0.7, 0.8, and 0.9, \method outperforms the original LLM, yielding an absolute performance increase of 1.3 to 2.6 points. These findings highlight \method's potential to enhance the reasoning capabilities of LLMs within the same length budget.

\begin{figure*}[t]
\centering
\includegraphics[width=0.95\textwidth]{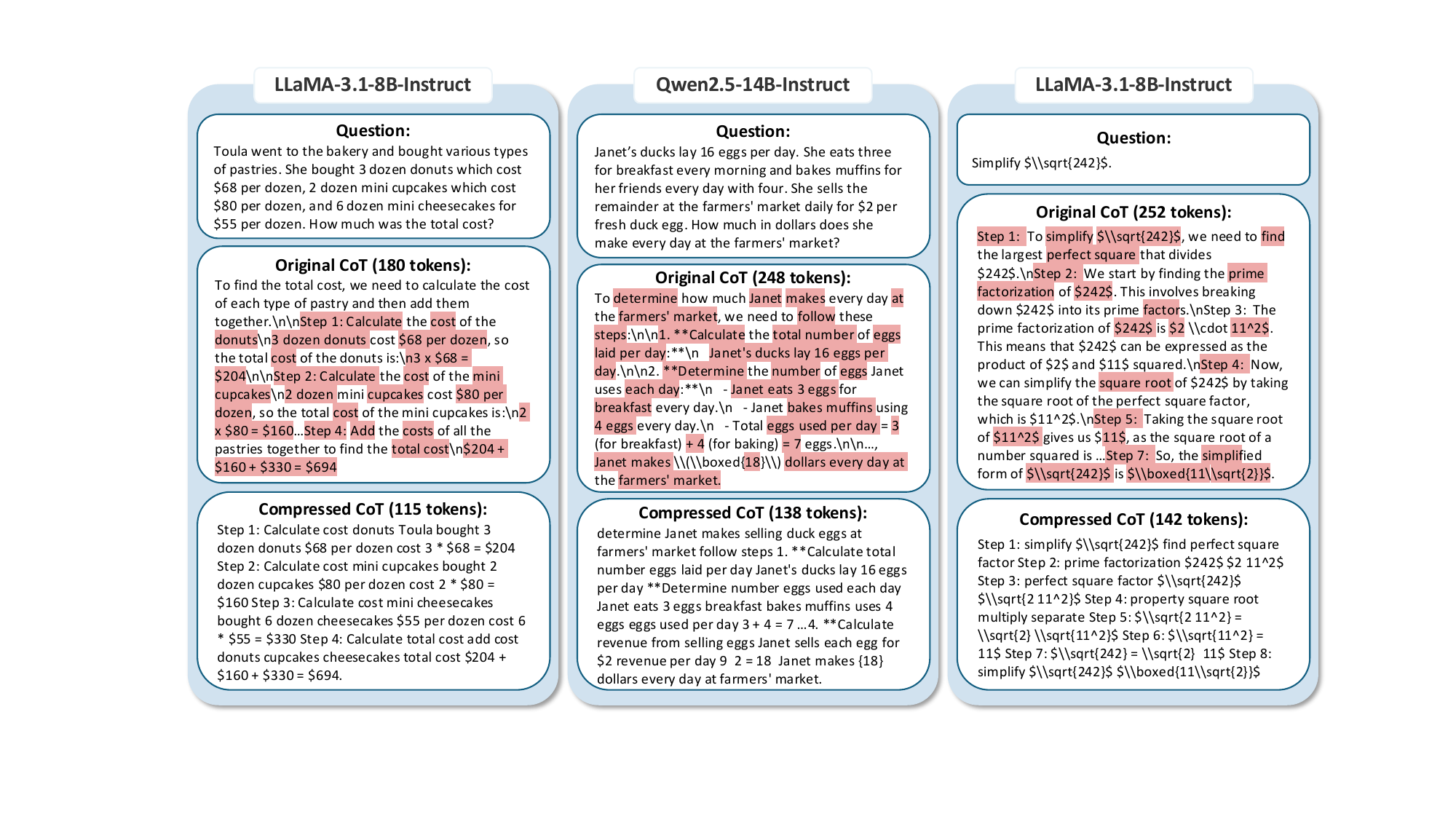}
\caption{Three CoT compression examples from \method. For each sample, we list the question, original CoT outputs from corresponding LLMs, and the compressed CoT by \method. The tokens that appear in both the original CoT and the compressed CoT are highlighted in \sethlcolor{pink}\hl{red}.}
\label{fig:cases}
\end{figure*}

\paragraph{Case Study} 
Figure~\ref{fig:cases} presents several examples of \method, derived from the test sets of GSM8K and MATH-500. These examples clearly illustrate that \method allows LLMs to learn shortcuts between critical reasoning tokens, rather than generating shorter CoTs from scratch. For instance, in the first case, \method facilitates LLaMA-3.1-8B-Instruct to skip semantic connectors such as ``\textit{of}'' and ``\textit{the}'', as well as expressions that contribute minimally to the reasoning, such as the first sentence. Notably, we observe that numeric values and mathematical equations are prioritized for retention in most cases. This finding aligns with recent research~\cite{Ma:2024mathmatters}, which suggests that mathematical expressions may contribute more significantly to reasoning than CoT in natural language. Furthermore, we find that \method does not reduce the number of reasoning steps but instead trims redundant tokens within those steps.
\section{Related Work}

\paragraph{Efficient CoT} While Chain-of-Thought (CoT) enhances the reasoning performance of LLMs, it introduces significant computational overhead. Researchers have sought methods to reduce this overhead while retaining the benefits of CoT. One intuitive approach is to simplify~\cite{marconato2024not}, skip~\cite{Ding:2024cotshortcut, liu2024skipstep}, or generate reasoning steps in parallel~\cite{ning2023skeleton}. Another research direction involves compressing CoTs into latent representations~\cite{goyal2024think, deng2024explicit, hao2024training, cheng2024compressed}, allowing LLMs to reason without explicitly generating discrete tokens. To mitigate CoT redundancy, \citet{han2024token} guides token consumption through dynamic token budget estimation. \citet{kang2024c3ot} prompts GPT-4 to shorten CoT trajectories, and then fine-tunes LLMs using compressed CoTs. In contrast, this work focuses on pruning CoT tokens based on their semantic importance. Moreover, \method leverages a small LM for token pruning, significantly reducing computational overhead.

\paragraph{Prompt Compression} The growing demand for long-context prompts has led to substantial computational and memory challenges. To address this, researchers have explored various prompt compression techniques. One intuitive approach involves using a lightweight LM to generate more concise prompts~\cite{chuang-etal-2024-learning}. Considering that natural language formats inevitably contain redundancy, some studies have introduced implicit continuous tokens to represent long-context inputs~\cite{chevalier-etal-2023-adapting, ge2024incontext, mohtashami2023randomaccess}. Another line of research focuses on directly compressing prompts by filtering low-informative tokens~\cite{li:2023selective, jiang2023:llmlingua, pan:2024llmlingua2}. For instance, Selective Context uses the perplexity of LLMs to measure token importance and removes less important tokens. LLMLingua-2~\cite{pan:2024llmlingua2} introduces a small bidirectional language model for token importance measurement and trains this LM with \texttt{GPT-4} compression data, which serves as the token importance metric in this work.
\section{Conclusion}
\label{sec:conclusion}

This work introduces \method, a simple yet effective approach for controllable Chain-of-Thought (CoT) compression. \method is built upon the semantic importance of CoT tokens --- By selectively skipping less important tokens while preserving critical ones, \method enables LLMs to generate compressed CoTs with adjustable ratios, thereby striking an expected balance between reasoning efficiency and accuracy. Extensive experiments across various LLMs and tasks validate the effectiveness of \method. We hope our investigations in \textit{token skipping} will offer valuable insights for advancing efficient CoT research and inspire future studies in this area.
\section*{Limitations}
\label{subsec:limitation} 

Due to computational constraints, experiments with larger LLMs, such as Qwen2.5-32B-Instruct and Qwen2.5-72B-Instruct, were not conducted. We believe that \method could achieve a more favorable trade-off between reasoning performance and CoT token usage on these models. Additionally, the token importance measurement used in our study, derived from the LLMLingua-2 compressor~\cite{pan:2024llmlingua2}, was not specifically trained on mathematical data. This limitation may affect the compression effectiveness, as the model is not optimized for handling numerical tokens and mathematical expressions. Furthermore, experiments with long-CoT LLMs, such as QwQ-32B-Preview, were also excluded due to computational constraints. We plan to explore these aspects in future work, as we anticipate that \method’s potential can be further realized in these contexts.

\section*{Acknowledgements}
We thank all anonymous reviewers for their insightful comments and valuable feedback during the review process. The work described in this paper was supported by Research Grants Council of Hong Kong (PolyU/15207122, PolyU/15209724, PolyU/15207821, PolyU/15213323) and PolyU internal grants (BDWP). 

\section*{Ethics Statement}
\label{subsec:ethics} 
The datasets used in our experiment are publicly released and labeled through interaction with humans in English. In this process, user privacy is protected, and no personal information is contained in the dataset. The scientific artifacts that we used are available for research with permissive licenses. And the use of these artifacts in this paper is consistent with their intended use. Therefore, we believe that our research work meets the ethics of ACL.

\bibliography{custom}

\clearpage

\appendix

\section*{Appendix}
\section{CoT Recovery}
\label{appendix:recovery}

In this section, we provide the detailed prompt for our recovery experiments, which is illustrated in Figure~\ref{fig:recovery-prompt}. Besides, we present the CoT recovery result from \texttt{GPT-4o}~\cite{gpt-4} in Figure~\ref{fig:recovery-gpt}. The recovered results demonstrate that \texttt{GPT-4o} could understand the compressed CoT content and correctly restore the original CoT process.

\begin{figure}[htbp]
\begin{tcolorbox}[colback=blue!5!white,colframe=blue!75!black,title=Input Prompt for CoT Recovery,fontupper=\footnotesize,fonttitle=\scriptsize]
Could you please recover the following compressed Chain-of-Thought output of a mathematical question to its original full content?

\vskip 0.1in

The question is: 

Marcus is half of Leo’s age and five years younger than Deanna. Deanna is 26. How old is Leo?

\vskip 0.1in

The compressed Chain-of-Thought: 

break down Deanna 26 Marcus five years younger 26 - 5 21 Marcus half Leo's age Leo twice Marcus's age Marcus 21, Leo's age 2 x 21 = 42.

\vskip 0.1in

Original Full Chain-of-Thought:
\end{tcolorbox}
\vskip -0.1in
\caption{Input prompt for LLaMA-3.1-8B-Instruct designed to recover the compressed CoT from a GSM8K math problem.}
\vskip -0.1in
\label{fig:recovery-prompt}
\end{figure}

\begin{figure}[htbp]
\begin{tcolorbox}[colback=blue!5!white,colframe=blue!75!black,title=Revovering the Compressed Chain-of-Thought,fontupper=\footnotesize,fonttitle=\scriptsize]
\textbf{Compressed CoT}: break down Deanna 26 Marcus five younger 26 - 5 21 Marcus half Leo's age twice Marcus Marcus 21, Leo's age 2 x 21 = 42.

\vskip 0.1in
        
\textbf{Recovered CoT}: 1. We know that Deanna is 26 years old. 2. Marcus is five years younger than Deanna. So, Marcus’s age is \( 26 - 5 = 21 \). 3. Marcus is also half of Leo’s age, which means Leo’s age is twice Marcus’s age. 4. Since Marcus is 21 years old, Leo’s age is \( 2 \times 21 = 42 \). So, Leo is 42 years old.
\end{tcolorbox}
\vskip -0.1in
\caption{Recovering the compressed CoT for GSM8K math word problem using GPT-4o.}
\vskip -0.1in
\label{fig:recovery-gpt}
\end{figure}

\section{Experimental Details}
\label{appendix:exp_details}

\subsection{Implementation Details}
\label{appendix:training_details}
We utilize LLMLingua-2~\cite{pan:2024llmlingua2} as the token importance metric to generate our compressed CoT training data. The compression ratio $\gamma$ is randomly selected from $\{0.5, 0.6, 0.7, 0.8, 0.9, 1.0\}$ for each training sample. We adopt LoRA~\cite{lora}, an efficient and reproducible approach that has been widely verified as effective in LLM fine-tuning, to train our models. The rank $r$ is set to 8, and the scaling parameter $\alpha$ is set to 16. We train the models for 3 epochs on both datasets. The peak learning rate is set to 5e-5, following a cosine decay schedule. We use AdamW~\cite{AdamW} for optimization, with a warmup ratio of 0.1. We implement our training process using the \texttt{LLaMA-Factory}~\cite{llamafactory} library. Inference for both our method and all baselines is performed using the Huggingface transformers package. During inference, the maximum number of tokens \texttt{max\_len} is set to 512 for GSM8K and 1024 for MATH\footnote{Since many samples reach the maximum length when testing \method on MATH-500, we adjust its length budget to \texttt{max\_len}$\times\gamma$, with no adjustment for GSM8K.}. All experiments are conducted using Pytorch 2.1.0 on 2$\times$NVIDIA GeForce RTX 3090 GPU (24GB) with CUDA 12.1, and an Intel(R) Xeon(R) Platinum 8370C CPU with 32 cores.

\subsection{Detailed Results with Qwen}
\label{appendix:qwen_detail}
We provide detailed experimental results of the Qwen2.5-Instruct series evaluated on GSM8K in Table~\ref{tab:qwen}. As the model scale increases, there is less performance degradation at higher compression ratios, indicating that larger LLMs are better at identifying shortcuts between critical reasoning tokens, enabling more efficient CoT generation.

\begin{table}[htbp]
\centering
\small
\setlength{\tabcolsep}{1.4mm}
\begin{tabular}{@{}llcrrc@{}}
\toprule
\textbf{Scale} &\textbf{Methods} &\textbf{Ratio} &Accuracy &Tokens &\textit{Act}Ratio \\ \midrule
\multirow{7}{*}{\texttt{3B}} & \texttt{Original}  & - & 83.7\cred{(0.0 \downarrow)} &314.87 &- \\ \cmidrule{2-6}
&\multirow{6}{*}{\method}  &1.0 & 83.4\cred{(0.3 \downarrow)} &318.79 & 1.00  \\
& &0.9 & 83.2\cred{(0.5 \downarrow)} &262.99 & 0.83\\
& &0.8 & 81.6\cred{(2.1 \downarrow)} &250.71 & 0.79\\
& &0.7 & 80.1\cred{(3.6 \downarrow)} &233.03 & 0.73\\
& &0.6 & 77.3\cred{(6.4 \downarrow)} &199.55 & 0.63\\
& &0.5 & 74.4\cred{(9.3 \downarrow)} &170.55 & 0.54 \\\midrule
\multirow{7}{*}{\texttt{7B}} & \texttt{Original}  & - & 91.4\cred{(0.0 \downarrow)} &297.83 &- \\ \cmidrule{2-6}
&\multirow{6}{*}{\method}  &1.0 & 91.7\cred{(0.3 \uparrow)} &295.78 & 1.00  \\
& &0.9 & 91.1\cred{(0.3 \downarrow)} &254.77 & 0.86\\
& &0.8 & 90.1\cred{(1.3 \downarrow)} &237.27 & 0.80\\
& &0.7 & 89.9\cred{(1.5 \downarrow)} &216.73 & 0.73\\
& &0.6 & 87.9\cred{(3.5 \downarrow)} &178.07 & 0.60\\
& &0.5 & 86.0\cred{(5.4 \downarrow)} &151.44 & 0.51 \\\midrule
\multirow{7}{*}{\texttt{14B}} & \texttt{Original}  & - & 93.1\cred{(0.0 \downarrow)} &313.11 &- \\ \cmidrule{2-6}
&\multirow{6}{*}{\method}  &1.0 & 93.0\cred{(0.1 \downarrow)} &314.55 & 1.00  \\
& &0.9 & 93.3\cred{(0.2 \uparrow)} &269.22 & 0.86\\
& &0.8 & 93.2\cred{(0.1 \uparrow)} &247.24 & 0.79\\
& &0.7 & 93.4\cred{(0.3 \uparrow)} &218.62 & 0.70\\
& &0.6 & 92.7\cred{(0.4 \downarrow)} &180.68 & 0.57\\
& &0.5 & 91.4\cred{(1.7 \downarrow)} &156.85 & 0.50 \\
\bottomrule
\end{tabular}
\caption{Experimental results on the Qwen2.5-Instruct series. We report accuracy, average CoT token count, and actual compression ratio (\textit{Act}Ratio) for comparison.}
\label{tab:qwen}
\end{table}

\subsection{Detailed Prompts}
\label{appendix:prompt_detail}
We demonstrate the detailed prompts used in our main experiments in Table~\ref{tab:prompts}.

\begin{table}[htbp]
\small
\centering
\setlength{\tabcolsep}{1.4mm}
\begin{tabular}{@{}>{\arraybackslash}m{55pt}>{\arraybackslash}m{150pt}@{}}
\toprule
\textbf{Methods} &\textbf{Detailed Prompts}  \\ \midrule
\texttt{BeConcise} & Be concise. \\
\texttt{OnlyNumbers} & Only use numbers or equations. \\
\texttt{AbbreWords} & Abbreviate words as much as possible. \\
\texttt{LC-Prompt} & Please reduce 50\% of the words in your Chain-of-Thought process.\\
\bottomrule
\end{tabular}
\caption{Details for prompt-based baselines.}
\label{tab:prompts}
\end{table}
\section{Additional Experiments}
\label{appendix:exp}

\subsection{Out-of-domain Evaluation}
\label{appendix:ood}
To assess the generalizability of \method beyond the training domain data, we conducted an additional out-of-domain evaluation. Specifically, we fine-tuned LLaMA-3.1-8B-Instruct on the MATH training data and evaluated \method on both the in-domain MATH-500 and two out-of-domain benchmarks, GSM8K and MMLU-STEM~\cite{hendrycks2021mmlu}. MMLU-STEM includes a diverse set of STEM subjects from the full MMLU dataset.

The results in Table~\ref{tab:ood} suggest that \method maintains strong generalizability on out-of-domain scenarios. The model adheres closely to specified compression ratios while preserving accuracy. Notably, on the MMLU-STEM test set, \method exhibits comparable performance to the original LLM with $\bm{40\%}$ token trimming. Even at a compression ratio of 0.5, the model maintains strong reasoning capabilities, with only $0.4\%$ absolute performance degradation.

\begin{table}[t]
\centering
\small
\setlength{\tabcolsep}{1.4mm}
\begin{tabular}{@{}lcrrc@{}}
\toprule
\textbf{Methods} &\textbf{Ratio} &Accuracy &Tokens &\textit{Act}Ratio \\ \midrule
\multicolumn{5}{c}{\textit{MATH-500 (in-domain)}} \\ \midrule
\texttt{Original}  & - & 48.6\cred{(0.0 \downarrow)} & 502.60 &- \\ \midrule
\multirow{6}{*}{\method} & 1.0 & 48.2\cred{(0.4 \downarrow)} &504.79 & 1.00  \\
&0.9 & 47.8\cred{(0.8 \downarrow)} & 448.31 & 0.89\\
&0.8 & 47.3\cred{(1.3 \downarrow)} & 398.94 & 0.79\\
&0.7 & 46.7\cred{(1.9 \downarrow)} & 349.13 & 0.69\\
&0.6 & 42.0\cred{(6.6 \downarrow)} & 318.36 & 0.63\\
&0.5 & 40.2\cred{(8.4 \downarrow)} & 292.17 & 0.58 \\\midrule
\multicolumn{5}{c}{\textit{GSM8K (out-of-domain)}} \\ \midrule
\texttt{Original}  & - & 86.2\cred{(0.0 \downarrow)} &213.17 &- \\ \midrule
\multirow{6}{*}{\method} & 1.0 & 86.0\cred{(0.2 \downarrow)} &214.49 & 1.00  \\
&0.9 & 84.9\cred{(1.3 \downarrow)} &201.84 & 0.95\\
&0.8 & 83.7\cred{(2.5 \downarrow)} &175.24 & 0.82\\
&0.7 & 82.6\cred{(3.6 \downarrow)} &152.32 & 0.71\\
&0.6 & 79.8\cred{(6.4 \downarrow)} &136.95 & 0.64\\
&0.5 & 76.6\cred{(9.6 \downarrow)} &122.55 & 0.58 \\\midrule
\multicolumn{5}{c}{\textit{MMLU-STEM (out-of-domain)}} \\ \midrule
\texttt{Original}  & - & 58.5\cred{(0.0 \downarrow)} &356.31 &- \\ \midrule
\multirow{6}{*}{\method} & 1.0 & 58.4\cred{(0.1 \downarrow)} &354.25 & 1.00  \\
&0.9 & 59.4\cred{(0.9 \uparrow)} &327.18 & 0.92\\
&0.8 & 59.3\cred{(0.8 \uparrow)} &286.15 & 0.80\\
&0.7 & 58.9\cred{(0.4 \uparrow)} &257.26 & 0.72\\
&0.6 & 59.2\cred{(0.7 \uparrow)} &225.33 & 0.63\\
&0.5 & 58.1\cred{(0.4 \downarrow)} &188.87 & 0.53 \\
\bottomrule
\end{tabular}
\caption{Out-of-domain results on LLaMA-3.1-8B-Instruct. We report accuracy, average CoT token count, and actual compression ratio (\textit{Act}Ratio) for comparison.}
\label{tab:ood}
\end{table}

\subsection{Evaluation Beyond Math}
\label{appendix:eval-qa}
To demonstrate the generalizability of \method beyond mathematical reasoning, we present results on CommonsenseQA~\cite{commonsenseqa}, a widely used multiple-choice question answering dataset that requires diverse commonsense knowledge to predict correct answers. For this experiment, we used 9,700 samples from the training set and evaluated \method on the validation set.

\begin{table}[t]
\centering
\small
\setlength{\tabcolsep}{1.4mm}
\begin{tabular}{@{}lcrrc@{}}
\toprule
\textbf{Methods} &\textbf{Ratio} &Accuracy &Tokens &\textit{Act}Ratio \\ \midrule
\multicolumn{5}{c}{\textit{Qwen2.5-7B-Instruct}} \\ \midrule
\texttt{Original}  & - & 80.3\cred{(0.0 \downarrow)} & 272.13 &- \\ \midrule
\multirow{6}{*}{\method} & 1.0 & 80.4\cred{(0.1 \uparrow)} &273.64 & 1.00  \\
&0.9 & 80.9\cred{(0.6 \uparrow)} & 245.70 & 0.90\\
&0.8 & 81.1\cred{(0.8 \uparrow)} & 218.73 & 0.80\\
&0.7 & 82.0\cred{(1.7 \uparrow)} & 188.78 & 0.69\\
&0.6 & 81.5\cred{(1.2 \uparrow)} & 153.17 & 0.56\\
&0.5 & 80.6\cred{(0.3 \uparrow)} & 128.43 & 0.47 \\\midrule
\multicolumn{5}{c}{\textit{Qwen2.5-14B-Instruct}} \\ \midrule
\texttt{Original}  & - & 82.1\cred{(0.0 \downarrow)} &247.81 &- \\ \midrule
\multirow{6}{*}{\method} & 1.0 & 83.8\cred{(1.7 \uparrow)} &247.34 & 1.00  \\
&0.9 & 82.9\cred{(0.8 \uparrow)} &221.75 & 0.95\\
&0.8 & 82.3\cred{(0.2 \uparrow)} &199.07 & 0.82\\
&0.7 & 82.1\cred{(0.0 \downarrow)} &172.44 & 0.71\\
&0.6 & 82.0\cred{(0.1 \downarrow)} &146.68 & 0.59\\
&0.5 & 82.1\cred{(0.0 \downarrow)} &121.03 & 0.49 \\
\bottomrule
\end{tabular}
\caption{Experimental results on CommonsenseQA with Qwen2.5-Instruct models. We report accuracy, average CoT token count, and actual compression ratio (\textit{Act}Ratio) for comparison.}
\label{tab:qa}
\end{table}

Experimental results on Qwen2.5-Instruct models are shown in Table~\ref{tab:qa}, which demonstrate that \method effectively reduces CoT length by 50\% without any performance degradation. These findings further highlight the generalizability of \method beyond the mathematical reasoning.

\end{document}